\documentclass{article}
\pdfoutput=1
\usepackage{PRIMEarxiv}

\usepackage[utf8]{inputenc} 
\usepackage[T1]{fontenc}    
\usepackage{hyperref}       
\usepackage{xurl}            
\usepackage{booktabs}       
\usepackage{amsfonts}       
\usepackage{amsmath}        
\usepackage{amssymb}        
\usepackage{nicefrac}       
\usepackage{microtype}      
\usepackage{fancyhdr}       
\usepackage{graphicx}
\usepackage{array}          
\providecommand{\tightlist}{\setlength{\itemsep}{0pt}\setlength{\parskip}{0pt}}
\graphicspath{{./figures/}{./}}

\title{Get In, Decide, Prove:\\ A Merchant-Side Readiness Benchmark for Agentic Commerce}

\author{
    Adrian de Valois-Franklin \\
    Evolutionairy AI \\
    Toronto, Canada \\
    \And
    Alex Bogdan \\
    Evolutionairy AI \\
    Toronto, Canada \\
    \And
    Anas Ibrahim \\
    Evolutionairy AI \\
    Toronto, Canada \\
    \AND
    Ibrahim Shaer \\
    Evolutionairy AI \\
    Toronto, Canada \\
    \And
    Tamaz Andguladze \\
    Evolutionairy AI \\
    Toronto, Canada \\
  }

\begin{document}
\maketitle

\begin{abstract}
AI agents now browse and buy on people's behalf, and benchmarks measure how well the agents perform. Far less is known about the other side of the transaction: whether a merchant's website lets an agent in, publishes terms precise enough to decide a purchase, and leaves evidence of what was agreed. This paper presents a merchant-side benchmark built on those three questions and reports a pilot of it. The central measurement, decision sufficiency, rates each criterion of a concrete buyer mandate (such as total cost, cancellation or return terms, service date, and variant identity) as determinate, indeterminate, or unknown from the merchant's own published evidence. Readiness scores are reported with their evidence coverage and missingness bounds; access is reported separately for a declared bot, a declared headless browser, and a user-present browser; and each observed term is mapped to the RAILS admissibility class of the evidence that could support it in a dispute. Four large merchants received the full assessment. Three refused the declared profiles at their offer pages. No mandate could be confirmed as met from the published terms: each merchant left a total or a refund term undetermined, and at one merchant the terms that were stated settled the decision as a rejection. An access census of 555 sites of merchants that two payment processors publicly name as clients found the home page unreadable to both declared profiles at 105 sites (18.9\%), from 8\% of sports and venue sites to 29\% of retail sites and 43\% of the 35 travel sites. A declared assessment of 330 merchants with an online store, scored by deterministic rules without a rater, found that evidence coverage rather than the score decides who can be ranked: 115 merchants (35\%) reached the ranking threshold under at least one declared profile, the rankable profiles scored a median of 89 to 90, and material terms was the one pillar well below the others, because the all-in total was decidable at 18 merchants. The pilot shows that the measurements can be taken. It does not validate them, it does not rank merchants, and the weights and the projected clearing band await validation against outcomes.
\end{abstract}

\keywords{agentic commerce \and AI agents \and merchant readiness \and benchmark \and decision sufficiency \and web measurement \and missing data \and bot management \and price transparency \and clearing \and RAILS}

\section{Introduction}
\label{sec:intro}

An agent asked to book a three-night stay at a resort, for no more than a fixed budget, reaches the resort's rate page in a few steps. The page lists a nightly rate, says that the rate includes a mandatory charge, and says nothing about tax. The agent has read everything the page shows for those dates, and it still cannot tell whether the stay fits the budget. If it books and the final charge exceeds the budget, the person who delegated the task will ask what the agent was shown. At the moment of decision, the only record of the displayed terms is the agent's own.

The case comes from the pilot reported in this paper, and it is not a failure of the agent. At the same resort, one term that was stated exactly, the cancellation rule, let the agent decline the offer without settling the total (Section~\ref{sec:deepdives}). Consumer agents that browse and buy on a person's behalf are now sold as products~\cite{openai_agent2025,muse2026}, and when one large assistant scaled back its in-chat checkout in March 2026, it returned the purchase to the merchant's own checkout~\cite{openai_discovery2026}. The merchant's website remains a place where agent purchases are decided and completed. The research community, meanwhile, has measured the agents closely. Web-agent benchmarks~\cite{zhou2023webarena,deng2023mind2web,he2024webvoyager,xue2025illusion,zhang2026clawbench} and their shopping successors~\cite{wang2025shoppingbench,lyu2025deepshop,peeters2025webmall,tou2025shoppingcomp,allouah2025aces} fix the task suite and vary the agent. The access side has been measured for crawlers~\cite{longpre2024consent,liu2024somesite,hoffmann2026aitxt}. Industry, for its part, has produced a large number of readiness checkers, which mostly audit markup, protocol files, and crawler access~\cite{cloudflare2026readiness,shopify2026readiness,serviloq2026}; one runs a staged purchase journey against real stores~\cite{ottom2026}. None of these rates the condition in the opening case, a page that is reachable, readable, and insufficient, criterion by criterion against a stated mandate; the evaluation stage of the journey benchmark comes closest.

That condition matters most where money is contested. Card-network rules now address purchases made by agents, and mandate protocols record what the person authorized~\cite{visa_rules2026,mastercard_vi2026,parikh2025ap2}. A dispute over an agent's purchase can therefore turn on what the agent was shown and what it accepted. The protocols sign the transaction and not the decision that led to it~\cite{louck2026signing}, and a systematization of blockchain agent-to-agent payments lists weak intent binding and limited accountability among the open problems~\cite{zhang2026sokpayments}. Whether a purchase can later be cleared (that is, whether anyone can determine that the agent met its delegated obligation~\cite{rails2026}) depends in part on what the merchant published before the purchase.

This paper describes a benchmark that holds the agent's task fixed and varies the merchant, and it reports a pilot. The benchmark asks three questions of a merchant's website. Can an agent \emph{get in}: reach the offer and its terms through a route the merchant permits? Can it \emph{decide}: do the merchant's own statements determine whether the offer satisfies a concrete buyer mandate? And could anyone \emph{prove}, afterward, what the agent agreed to? The paper makes five contributions.

\begin{enumerate}
\tightlist
\item \textbf{Decision sufficiency.} Each criterion of a concrete buyer mandate (the all-in total, the cancellation or return terms, the delivery or service date, the identity of the variant) is rated determinate, indeterminate, or unknown from the merchant's own published evidence, and a task is sufficient only when every criterion is determinate (Section~\ref{sec:sufficiency}).
\item \textbf{A score that carries its missing evidence.} Every readiness score is reported with its evidence coverage and with missingness bounds, and ranked placement is withheld below a coverage threshold (Section~\ref{sec:scoring}).
\item \textbf{Access by agent profile.} Results are reported separately for a declared bot, a declared headless browser, and a user-present browser; the declared profiles record every barrier as a result and never bypass it (Section~\ref{sec:benchmark}).
\item \textbf{A link from the website to clearing.} The RAILS admissibility classes and Clearing Passport~\cite{rails2026} are applied to merchant websites: each observed term is mapped to the class of evidence that could support it, a clearing band is projected from the result as an illustration until outcome data exist, and a per-merchant record is seeded whose outcome fields are reserved for processor data (Sections~\ref{sec:evidence} and~\ref{sec:passport}).
\item \textbf{A pilot.} Four large merchants received the full assessment, and 555 sites of merchants that two payment processors publicly name as clients received the access measurements (Section~\ref{sec:pilot}).
\end{enumerate}

Each of the first four has a partial precedent, which Section~\ref{sec:related} names. In a search of about 70 queries and 60 fetched sources up to 30 September 2026, we found no public benchmark or tool that combines them. The paper makes no claim beyond that. It does not rank merchants, it does not show that the measured gaps cause disputes or losses, and its weights and its band rule are design choices awaiting validation (Section~\ref{sec:limitations}).

\section{Background and Related Work}
\label{sec:related}

\subsection{Benchmarks that evaluate the agent}

WebShop~\cite{yao2022webshop}, WebArena~\cite{zhou2023webarena}, Mind2Web~\cite{deng2023mind2web}, and REAL~\cite{garg2025real} established the practice of scoring an agent against tasks on simulated, replicated, or recorded websites. WebVoyager~\cite{he2024webvoyager}, Online-Mind2Web~\cite{xue2025illusion}, and ClawBench~\cite{zhang2026clawbench} moved the tasks onto live sites, where the later studies find that agents complete far fewer of them than earlier results suggested. The commerce benchmarks follow the same design: ShoppingBench~\cite{wang2025shoppingbench} and WebMall~\cite{peeters2025webmall} vary the agent against a fixed catalog, DeepShop~\cite{lyu2025deepshop} and ShoppingComp~\cite{tou2025shoppingcomp} test it on live shops and open-world products, and market simulations such as ACES~\cite{allouah2025aces} and Magentic Marketplace~\cite{bansal2025magentic} study what agents choose and how markets of agents behave. A related line shows that manipulative interface designs steer agents~\cite{ersoy2025darkpatterns}, and steer them more readily than they steer people~\cite{cuvin2025decepticon}.

In all of this work, the website is the instrument and the agent is the subject. When the tasks run on live sites, a failure mixes what the agent could not do with what the site did not allow or did not say. A merchant-side benchmark separates the two by fixing the task and the collection method and letting the merchant vary.

\subsection{Who is allowed in}

Measurement of website-side restrictions has concentrated on crawlers that collect training data. Longpre et al.\ document the rapid growth of restrictions in \texttt{robots.txt} and terms of service~\cite{longpre2024consent}; Liu et al.\ find strong demand for such controls among content creators and little deployment of the stronger network-level blockers~\cite{liu2024somesite}; Kim et al.\ show that stricter directives are obeyed less~\cite{kim2025scrapers}; and Hoffmann et al.\ map the newer permission files, including \texttt{llms.txt}~\cite{howard2024llmstxt}, across about four million domains~\cite{hoffmann2026aitxt}. Network-scale data show crawling triggered by a user's action growing more than fifteenfold during 2025, while AI crawlers were the user agents most often fully disallowed in \texttt{robots.txt}~\cite{belson2025radar}; one large network now blocks bots it classifies as agents, by default, on pages of new domains that display ads~\cite{cloudflare2026aitraffic}. The legal position is unsettled as well. In August 2026, a federal appeals court vacated a preliminary injunction that a retailer had obtained against a shopping agent operating inside its users' accounts; the dispute centered on the developer's decision not to use a user-agent string that would announce an active AI agent~\cite{amazon_perplexity2026}.

A second line studies detection: browsing agents are distinguishable by behavior~\cite{wang2026fpagent}, and challenge-based defenses do not reliably stop them~\cite{ousat2026brokengates}. Proposals to repair the relationship include signed requests for automated clients~\cite{meunier2026webbotauth}, permission manifests~\cite{marro2025permission}, machine-readable access terms with signed receipts~\cite{chowdhury2026termstxt}, and interfaces built for agents~\cite{lu2025buildweb}; controlled experiments show that the interface a shop exposes changes agent outcomes~\cite{steiner2025mcpnlweb}. Practitioner tools have begun to compare access by identity: one open-source auditor fetches a store's pages under several declared AI user agents and compares each with a browser fetch~\cite{agentready2026}.

The cited studies measure declared policy, crawler compliance, or the detectability of agents. The benchmark's profiles ask a narrower question, whether a merchant serves an automated client that names itself and its operator, and they differ from security-oriented tests in one respect that matters for ethics and for interpretation: no declared profile hides that it is automated, no profile forges an identity or solves a challenge, and the user-present profile sends the person's ordinary browser user agent.

\subsection{Readiness checkers and provider-side measures}

Scoring a website for its readiness for agents is not new, and the benchmark should not be read as the first of its kind. Cloudflare's Agent Readiness score audits discoverability, content negotiation, bot access control, and protocol capabilities, and checks commerce protocols without counting them~\cite{cloudflare2026readiness}. Chrome's Lighthouse has an experimental agentic-browsing category~\cite{chrome2026lighthouse}. Commerce-specific checkers audit a product page's structured data and crawler access~\cite{shopify2026readiness}, and in one case policy information and checkout accessibility as well~\cite{serviloq2026}. Ottom comes closest in form. It runs its own declared agent through discover, evaluate, compare, and buy stages on real stores, its evaluation stage asks whether the store's own data answer what the shopper asked for, it ends a scan at bot protection without bypassing it, and it publishes a category only when at least 20 stores have been measured at every stage~\cite{ottom2026}. On the payments side, Ballerine offers readiness governance, with per-merchant readiness profiles, to payment service providers across their merchant base~\cite{ballerine2026}.

Several of these tools already handle thin evidence in some way. Ottom counts an unmeasured stage as zero, which caps the attainable score. IsAgentReady reports the coverage of its core checks beside the score~\cite{isagentready2026}. An index of commerce platforms withholds numeric scores from its lower tiers because the evidence would not support the precision~\cite{elogic2026}. One open-source auditor rescales the remaining weights when a check is skipped~\cite{agentready2026}. The observed score of Section~\ref{sec:scoring} is the same ratio; the difference is that a skipped test stays in the frame, where the coverage and the bounds show it, and that no rank is given below a coverage threshold.

Academic work on the provider side is recent. Elnaffar and Rashidi define an agent-ready website along interpretability, executability, and decision reliability, and test the design on one prototype shop~\cite{elnaffar2026agentready}; their decision reliability is the nearest conceptual precedent for decision sufficiency. Peng scores software products for the share of their functionality that agents can call~\cite{peng2026acfc}. RADAR separates whether an AI system can describe a public service from whether an agent can reach it, across 166 countries~\cite{jordan2026radar}. Wadi and Ma show that when reading a storefront's attributes is costly and the goal is vague, agents skip the attributes they need and make worse choices, which the authors attribute to the storefront's information architecture~\cite{wadi2026shopping}. For scoring under missing evidence, the closest precedent is outside commerce: Gao and Zhou label each outcome of an agent benchmark as passing, failing, or unknown on the stored evidence and report score bounds over the unknown cases~\cite{gao2026bounds}, which is the device the benchmark applies to merchants and which descends from partial identification with missing outcomes~\cite{manski2003partial,horowitz2000missing}.

Against this background, the benchmark's distinguishing features are narrower than ``readiness.'' Decision sufficiency rates each criterion of a stated mandate on a three-valued scale and keeps ``read and insufficient'' apart from ``not read.'' The score is reported per profile, with coverage, bounds, and a threshold below which no rank is given. To our knowledge, none of the tools or papers above grades the evidence that would remain of the terms an agent accepted.

\subsection{Mandates, protocols, and evidence}

Agent payment protocols bind a person's authorization to a bounded action. Google's Agent Payments Protocol introduces cryptographically signed Intent and Cart Mandates~\cite{parikh2025ap2}, and its specification requires the merchant to sign the checkout it offers and describes how mandates and receipts are brought together in a dispute~\cite{ap2spec2026}. The Agentic Commerce Protocol~\cite{acp2026} and the Universal Commerce Protocol~\cite{ucp2026} standardize checkout between agents and businesses, and the card networks add agent identification~\cite{visa_tap2025} and a record of what the shopper authorized~\cite{mastercard_vi2026}. These mechanisms create strong evidence for the facts they sign. They do not show that the signed cart matched the person's request~\cite{louck2026signing}, and they operate only where a merchant has adopted them.

RAILS~\cite{rails2026} specifies clearing for agent obligations: an Obligation Object states the task and its acceptance criteria, evidence is graded by admissibility class, and no financially material settlement may rest on evidence below the obligation's floor. The present paper takes those constructs as given and asks a prior, empirical question: what evidence of the offer's terms does a merchant's website make available today, and how far is it from what clearing would need?

\subsection{Price transparency}

The terms that decision sufficiency examines overlap with a concern of consumer law and of pricing research. Rules on total-price disclosure now exist in several jurisdictions; in the United States, for example, a federal rule has required since May 2025 that any price offered, displayed, or advertised for short-term lodging or live-event tickets disclose the total price~\cite{ftc2025fees}. That total may exclude government charges and shipping, which must be disclosed before the consumer consents to pay; a page that states that tax is added at checkout can therefore comply with the rule and still leave an agent's budget unchecked. Field experiments show why the total matters to human buyers: fees revealed late raise both the quantity and the quality purchased~\cite{blake2021salience}, and buyers tend to keep a choice made on a low base price after the full price appears~\cite{santana2020drip}. An agent with a hard budget is a different kind of buyer, since it can be instructed to stop when a total cannot be bounded. For that buyer, a missing tax basis is not a nudge but a gap in the evidence. The benchmark's cost metrics describe what an agent could observe. They are not legal conclusions, and this paper makes no finding on any merchant's compliance with any rule.

\section{The Benchmark}
\label{sec:benchmark}

Consider an instruction a person might give a shopping agent: book a round trip for one adult on two named dates, with one carry-on bag, for no more than \$200 in total, refundable in full if canceled within 24 hours. The instruction names a product, a budget, and a condition of recourse. Whether an agent can carry it out depends on three things the merchant controls: whether the agent is allowed to read the pages, whether the pages state the total and the refund terms precisely enough to be checked against the instruction, and whether anything would later show what the pages said. The benchmark measures those three things, and it reports them separately because they fail separately.

\subsection{Unit of assessment}

The unit of assessment is not a website. It is a triple: a merchant, a buyer mandate, and an agent profile. The same merchant can be fully readable to one kind of agent and closed to another, and the same page can settle one mandate and leave a second undecided. Every result is therefore recorded with the route taken, the merchant or location, the locale, currency and destination, the login state, the time of observation, the page sample, and the scanner version. A marketing domain often hands commerce to a separately operated booking or ordering domain; the operator responsible for each observation is recorded.

\begin{figure}
\centering
\includegraphics[width=\linewidth,keepaspectratio]{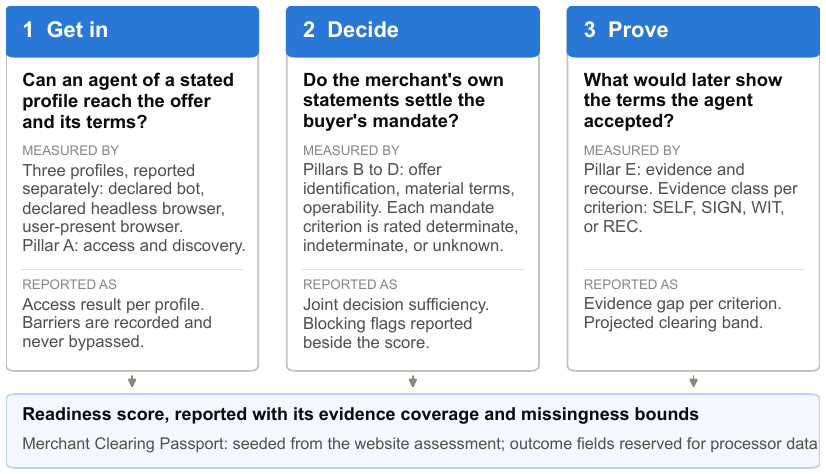}
\caption{The three questions of the benchmark and what each produces. Access is measured per agent profile; decision sufficiency is measured against a concrete buyer mandate; the evidence class of each term is reported as it stands before payment and as it would stand with merchant-signed offers, an independent witness, and processor records. The readiness score carries its evidence coverage and missingness bounds. The clearing band is an illustrative projection and is not calibrated against outcomes.}
\label{fig:framework}
\end{figure}

The three questions map onto three groups of measurements. \emph{Get in} asks whether an agent of a stated profile can reach the offer and its terms through a route the merchant permits. \emph{Decide} asks whether, once the pages are read, the merchant's own statements determine that the offer does or does not satisfy the mandate. \emph{Prove} asks what evidence would exist, after a purchase, of the terms the agent accepted, and how much weight that evidence could bear in a dispute. Figure~\ref{fig:framework} shows how the measurements connect.

\subsection{Assessment profiles}

An agent that announces itself and an agent that arrives inside a person's ordinary browser can be treated differently by the same site, so the benchmark fixes three profiles and reports them separately (Table~\ref{tab:profiles}).

\begin{table}[h]
\centering
\caption{Assessment profiles. Results are reported per profile and do not transfer between profiles.}
\label{tab:profiles}
\begin{tabular}{>{\raggedright\arraybackslash}p{3.2cm}>{\raggedright\arraybackslash}p{6.4cm}>{\raggedright\arraybackslash}p{5.6cm}}
\toprule
\textbf{Profile} & \textbf{Implementation} & \textbf{What it represents} \\
\midrule
Declared bot & HTTP client with an identifiable user agent that names the probe and links to the operator & A crawler or protocol client that identifies itself \\
Declared headless browser & Headless Chromium with the same token appended to its user agent, and no measure that conceals automation & A browser agent that renders pages without hiding that it is automated \\
User-present browser & A browser on the person's own device, operated by an AI browser assistant with the person present and approving each step; it sends the browser's ordinary user agent and performs no login, hold, cart, or payment & A consumer browser agent acting for a present user \\
\bottomrule
\end{tabular}
\end{table}

Three rules govern the profiles. A barrier met by a declared profile is an access result for that profile, recorded with the barrier's vendor where it can be identified, and it is never bypassed. A profile that could not run because of the assessment's own tooling or policy is NOT ASSESSED, which says nothing about the merchant. The user-present profile requires the person's approval for the specific merchant and for each step taken there; it measures what a user-present agent would see, and it is not a route for bulk collection past a barrier.

\subsection{Readiness metrics}

Twenty metrics in five pillars form the readiness score (Table~\ref{tab:metrics}; Appendix~\ref{app:metrics} gives the pass condition of each). Pillar A covers the first question, pillars B to D the second, and pillar E the third. Each metric is the proportion of its predefined eligible tests that pass. The pillar weights are initial engineering choices and not fitted predictors of any business outcome; Section~\ref{sec:limitations} describes how they would be tested.

\begin{table}[h]
\centering
\caption{The twenty readiness metrics. Weights are initial design choices.}
\label{tab:metrics}
\begin{tabular}{>{\raggedright\arraybackslash}p{4.2cm}p{1.2cm}>{\raggedright\arraybackslash}p{9.8cm}}
\toprule
\textbf{Pillar} & \textbf{Weight} & \textbf{Metrics} \\
\midrule
A. Access and discovery & 15\% & A1 readable route; A2 offer discovery; A3 offer content exposure; A4 navigation resolves \\
B. Offer identification & 25\% & B1 offer identity; B2 configuration specificity; B3 price interpretation; B4 availability \\
C. Material terms & 25\% & C1 cost components disclosed; C2 total cost determinable; C3 delivery or service determinacy; C4 return or cancellation determinacy \\
D. Observable operability & 20\% & D1 action target identifiable; D2 selection state visible; D3 prerequisites visible; D4 commerce handoff resolves \\
E. Evidence and recourse & 15\% & E1 counterparty disclosed; E2 policy applies to this offer; E3 offer referenceable; E4 recourse instructions \\
\bottomrule
\end{tabular}
\end{table}

The metrics measure clarity independently of generosity: a plainly stated ``no refunds'' passes C4, even though the offer is unsuitable for a buyer who requires a refund. Disclosure, decision sufficiency, and suitability are kept as separate fields. The operability metrics, for their part, inspect public page state and navigable links; they do not establish that adding to a cart, changing dates, or submitting payment will work. Those steps belong to a separately authorized journey test, because apparently harmless actions can reserve scarce inventory, create customer records, or trigger messages.

\subsection{Scoring without concealing missing evidence}
\label{sec:scoring}

A readiness score computed only over the pages an agent managed to read rewards a merchant for hiding the rest, and the observed score alone does so. The scoring rule therefore never shows that score without its coverage and its bounds, and it withholds a rank when coverage is low: it exposes what is hidden rather than preventing it.

The applicable atomic tests are fixed before inspection. Let $J$ be that set, with weights $a_j > 0$ and $\sum_{j \in J} a_j = 1$; each metric receives one quarter of its pillar's weight, distributed across its planned tests. Every test receives one of three statuses. A confirmed omission after the prescribed search is FAIL. An inaccessible page, an unresolved interpretation, or a scanner failure is UNKNOWN; a barrier therefore counts as FAIL for the access tests it blocks and as UNKNOWN for the content it hides. A conflict between two applicable statements is not an unresolved interpretation: the pages were read and do not settle the term, so the determinacy test is FAIL and the conflict is reported as a flag beside the score. A test is removed as not applicable only under a versioned vertical rule, never to improve a score. With $P$, $F$, and $U$ the sets of passed, failed, and unknown tests, the assessment reports
\begin{align*}
C &= \sum_{j \in P \cup F} a_j  && \text{(evidence coverage)} \\
L &= \sum_{j \in P} a_j          && \text{(earned proportion)} \\
S &= 100\, L / C, \quad C > 0    && \text{(observed score)} \\
{}[S_{\min}, S_{\max}] &= [\,100\,L,\; 100\,(L + 1 - C)\,] && \text{(missingness bounds)}
\end{align*}
The bounds give the score the merchant would receive if every unknown test failed and if every unknown test passed. They are worst-case bounds on missing information, in the sense of the partial-identification literature on missing outcomes~\cite{manski2003partial}; they are not confidence intervals and not a probability of a successful purchase.

Two profiles from the pilot of Section~\ref{sec:pilot} show why the score cannot travel alone. One merchant's declared profile and another's both received an observed score of 50.0. The first rested on coverage of 15\%, because a bot barrier hid every offer and policy page, and its bounds were 7.5 to 92.5. The second rested on coverage of 90\%, and its bounds were 45 to 55. The two results share a number and almost nothing else.

Three further rules apply. A profile is rankable only when coverage is at least 90\% overall and at least 80\% in every pillar; below that, it is published as an incomplete assessment, its measurements are retained, and ranked placement is withheld. These thresholds are governance proposals to be tested for practicality, not constants. Certain task-level conditions are reported beside the score and cannot be averaged away: an unresolved identity of the selected variant, an unresolved total under a hard budget, conflicting applicable cancellation terms, required evidence that is inaccessible, and the absence of any permitted route past a required step. And the same fact is never credited twice: a structured-data field that satisfies several technical checks earns its points once, and protocol badges, \texttt{llms.txt} files, and payment logos are recorded as diagnostics outside the score.

\subsection{Decision sufficiency}
\label{sec:sufficiency}

The central measurement asks whether the evidence a merchant publishes is enough to determine that an offer satisfies a concrete buyer obligation. The question is narrower than whether the merchant appears trustworthy, and it can be answered from the page.

Each task is drafted as an Obligation Object in the sense of RAILS~\cite{rails2026}: a task statement, a conjunction of acceptance criteria $c_1, \dots, c_k$, and the evidence each criterion requires. The pilot used seven criteria per task: the identity of the item, the configuration or dates, the all-in total against a budget, availability, the cancellation or return condition, the operator of record as the site states it, and the recourse channel. None of its mandates carried a delivery-date criterion. Each criterion receives one of three states:
\begin{itemize}
\tightlist
\item \textbf{DETERMINATE}: the value, its scope, and the time to which it applies are interpretable from the merchant's own evidence, and no unresolved conflicting evidence applies. A criterion can be determinate and unmet.
\item \textbf{INDETERMINATE}: the page was read, but the value, scope, or time cannot be fixed.
\item \textbf{UNKNOWN}: the evidence could not be observed under the profile.
\end{itemize}
A task has \emph{joint decision sufficiency} when every criterion is determinate, and the benchmark reports the proportion of tasks for which that holds. Suitability is reported separately as confirmed suitable, confirmed unsuitable, or unresolved, because an explicitly non-refundable rate is easy to evaluate and still wrong for a buyer who needs a refund. The benchmark also reports the \emph{decisive rejection rate}, the share of tasks in which one determinate and unmet criterion already settles rejection; an agent does not need the full conjunction to decline an offer.

Determinacy is a property of the terms, not of their format. A return policy carried in structured data is present and machine-readable; it is determinate only when its deadline, fee, exceptions, and refund method settle the criterion. ``Returns within 30 days, exceptions apply'' is machine-readable and indeterminate. The benchmark therefore records whether a term is present in typed data as a diagnostic beside the criterion state, and the state is earned from the content alone.

Table~\ref{tab:determinacy} gives the determinacy rules, each with a case from the pilot that fails it, either as a mandate criterion or as the corresponding readiness metric.

\begin{table}[h]
\centering
\caption{Determinacy rules, with failing cases observed in the pilot (paraphrased).}
\label{tab:determinacy}
\begin{tabular}{>{\raggedright\arraybackslash}p{3.0cm}>{\raggedright\arraybackslash}p{5.9cm}>{\raggedright\arraybackslash}p{6.3cm}}
\toprule
\textbf{Criterion} & \textbf{Determinate when} & \textbf{Failing case from the pilot} \\
\midrule
All-in total within budget & Amount, currency, charging basis, and mandatory fees are stated for the selected configuration and dates, or a finite bound is & A ``from'' price with no unit, fee, or tax basis for the selected dates; a nightly rate that includes a mandatory charge and states no tax basis \\
Cancellation or return & Deadline, fee, exceptions, and refund method are stated for the selected rate or item before commitment & Terms deferred to the reservation confirmation; one page promising cancellation without penalty while another says certain charges are not refunded \\
Delivery or service date & A delivery window for the destination, or explicit service dates, is stated & No shipping or start-of-service estimate before an order flow that the permitted route may not enter \\
Variant identity & The price is bound to the exact variant, room type, or rate plan on the page, and structured data does not contradict it & A nightly price bound to a room type and to no rate plan; a recurring plan price while the required equipment depends on the service address \\
Total behind a disallowed step & The total, or a finite bound, is available before any step the permitted route may not enter & Hardware, shipping, and tax not stated before an order flow that \texttt{robots.txt} disallows \\
\bottomrule
\end{tabular}
\end{table}

For each task, the assessment also records the effort a supported decision required: pages opened, machine actions, elapsed time, retries, criteria left unresolved, and human interventions. Effort is reported beside sufficiency. A task that the agent abandoned early would otherwise make the merchant look efficient.

\subsection{Observed information reliability}

Four rates describe whether the information a merchant publishes agrees with itself, and they are reported outside the readiness score with their counts. \textbf{I1}, cross-representation agreement, is the share of price, currency, variant, and availability comparisons across the rendered page, structured data, and declared feeds that show no material conflict. \textbf{I2}, policy consistency, is the share of offer-to-policy comparisons with no unresolved conflict; for date-dependent offers, disclosures are compared across at least two date contexts, and a fee statement present in one and missing in the other is an unresolved conflict. \textbf{I3} and \textbf{I4} measure repeat extraction success and short-window coherence over scheduled observations. A first crawl can estimate I1 and I2; I3 and I4 need repeats. Agreement over two comparisons is weak evidence, and a site with a single representation is unassessed on I1, not perfect. The four rates describe published information and say nothing about delivery.

\subsection{Evidence classes: what could be proved afterward}
\label{sec:evidence}

The third question concerns a dispute that has not yet happened. If the agent buys and the person later says the refund terms were not as promised, what would show what the merchant displayed? RAILS grades evidence by admissibility class, in the partial order $\mathrm{SELF} \preceq \mathrm{SIGN} \preceq \{\mathrm{WIT}, \mathrm{REC}\} \preceq \mathrm{ATT} \preceq \mathrm{PROOF}$, in which a witness signature and an external receipt are deliberately incomparable; ATT is an attestation from a trusted execution environment or a similarly hardened runner, and PROOF a cryptographic proof of correctness~\cite{rails2026}. No evidence source in Table~\ref{tab:classes} reaches either. The benchmark maps each source of evidence about an offer to the class it could support and to the claim that class licenses (Table~\ref{tab:classes}). RAILS defines SIGN for the acting agent's own signature. The benchmark applies the same class to a merchant's signature over its own offer, where it likewise gives non-repudiation and not truth; the merchant-signed checkout of AP2 is an artifact of this kind~\cite{ap2spec2026}.

\begin{table}[h]
\centering
\caption{Evidence sources, the RAILS admissibility class each can support, and the claim that class licenses.}
\label{tab:classes}
\begin{tabular}{>{\raggedright\arraybackslash}p{6.2cm}p{1.4cm}>{\raggedright\arraybackslash}p{7.6cm}}
\toprule
\textbf{Evidence source} & \textbf{Class} & \textbf{Claim it can support} \\
\midrule
The agent's own record of what it saw & SELF & That the agent reports having seen it \\
A merchant-signed offer, checkout session, or confirmation & SIGN & That the merchant committed to the stated terms; non-repudiation, not truth \\
An independent capture of the displayed terms by a witness that meets the deployment's witness requirements & WIT & That the merchant displayed these terms at a stated time under a stated profile; never that the terms are true or were honored \\
The processor's authorization and settlement record & REC & Amount, merchant of record, time, and payment status \\
A carrier or fulfillment system scan & REC & A delivery or fulfillment event and its time \\
\bottomrule
\end{tabular}
\end{table}

The mapping is deliberately narrow. A report signed by the crawler attests to an observation under its collection method; the signature alone does not promote a web snapshot to WIT, REC, or ATT. A capture qualifies as WIT only for the display claim, only when the witness is independent of the parties, and only under a deployment that accepts that witness. A hash preserves what was captured; it does not certify a merchant's claim, a fulfillment, or the time of capture without a further trusted mechanism.

For each criterion, the assessment reports two classes: the best class obtainable before payment as the site stands, and the class obtainable when merchant-signed offers, an independent witness, and processor records are added. The difference between the two is the evidence that the merchant or the clearing layer would have to supply.

\subsection{Projected clearing band and the Merchant Clearing Passport}
\label{sec:passport}

RAILS maps an obligation's exposure score to one of four policy bands, which set the verifier set, the settlement timing, and the margin. Two of the six inputs to that score, intent ambiguity and irreversibility, depend on what the merchant discloses~\cite{rails2026}. The benchmark projects a band from the criterion states while holding the other four inputs (authority risk, value at risk, agent history, and domain sensitivity) fixed (Appendix~\ref{app:bands}).

The rule is an illustration for discussion. It must not change how any payment is treated until it has been validated against outcomes.

RAILS defines the Clearing Passport as a cross-transaction reliability record for agents, tools, providers, and verifiers. The benchmark extends it to merchants. A \emph{Merchant Clearing Passport} is seeded from the website assessment: readiness by profile, decision sufficiency, blocking flags, I1 to I4, and the projected band. It reserves outcome fields (agent-initiated transactions, disputes, refunds, and cleared obligations) that only processor data can fill. The seed describes observed information; the outcome fields, once validated, describe performance. Every report keeps the two apart.

\subsection{Collection rules}
\label{sec:collection}

The declared profiles follow rules that a merchant could verify from its own logs. They use an identifiable user agent: the HTTP client sends \texttt{RAILS-ReadinessProbe/0.1 (+https://evolutionairy.ai; research probe)}, and the headless browser appends \texttt{RAILS-ReadinessProbe/0.1 (+https://evolutionairy.ai)} to Chromium's own string. They read crawler directives with the group semantics of RFC~9309~\cite{rfc9309}, under which the longest matching rule wins and a \texttt{robots.txt} that is unreachable because of a server or network error is treated as a complete disallow. They wait a stated \texttt{Crawl-delay} between requests, up to a cap of 120 seconds. Requests to one site run in sequence at about one per second. Cart, checkout, booking, and login URLs are recorded and never requested. A barrier is recorded, with its vendor where identifiable, and is never bypassed: no challenge is solved, no identity is changed, and no request is retried in another guise. Each planned URL is requested once per profile, whether or not an earlier one was refused. A response that returns HTTP 200 with an ordinary web page on a protocol path is reported as a soft-404, since an agent can mistake it for a protocol answer, and a directive that does not match the live URL structure is reported as such. Robots directives guide crawlers and are not access authorization, which is why the scope of an assessment is set by agreement with the merchant or its processor wherever that is possible.

The user-present profile runs in the person's ordinary browser, operated by an AI browser assistant with the person present and approving each step. It submits no personal data, holds no inventory, opens no cart, and stops before any booking or order step. Where a site shows prices only after a date search, one read-only search is submitted with the person's explicit permission.

Every observation is stored as a record carrying the merchant, profile, scenario, metric and test identifiers, the planned weight, status and reason code, the time, locale, destination, and vantage (the network location the observation was made from), a source pointer, the exact excerpt, a capture hash with its scope, the extractor and rubric versions, and the origin of the signal, which states whether a hosted commerce platform supplies it by default or the merchant configured it (Appendix~\ref{app:record}). The vantage and origin fields were added after the pilot, which recorded neither; the pilot and the census were observed from Ontario, Canada. Readiness is scored on what an agent can observe, whoever put it there, and the origin field lets a report show how much of a merchant's result rests on platform defaults. Personal data and payment credentials are kept out of the records.

\section{Pilot Study}
\label{sec:pilot}

The pilot had three parts. Four merchants received the full assessment against one mandate each, to test whether the constructs can be applied and what they show. A larger set of sites then received the access measurements alone, to test whether the first finding of the four holds beyond them. And 330 merchants received the twenty metrics by rule under the two declared profiles, to test whether the measurements can be taken at scale without a rater and what then decides who can be ranked.

\subsection{Four full assessments}
\label{sec:deepdives}

\textbf{Setup.} The four merchants are an airline (M1), a resort (M2), a subscription service that ships hardware (M3), and a hotel property of a large brand (M4). All are large consumer brands. One of the two processors of Section~\ref{sec:census} names three of them as clients in its public materials; the fourth appears only among client logos on that processor's website, a weaker link that the census sample would not have admitted. They are not identified here, because each result rests on one observation and one rater, and details that would identify them are omitted or generalized. Each merchant received one mandate in locale en-US and one observation on 29 September 2026, made from a network location in Ontario, Canada. The mandates followed one template: a named product or room, fixed dates or a service address, a hard all-in budget, and a refund or cancellation condition. Where the declared profiles were refused at the offer pages (M1, M2, and M4), the user-present profile was also run. For M3, the declared headless browser read every public page, so the user-present profile was not run. The two declared profiles are reported separately where their observations differed. At M1, M2, and M4 they met the same barrier at every requested page, and one set of verdicts covers both; at M3 the declared headless browser read every public page while the plain HTTP client received an empty shell at each, so Table~\ref{tab:pilot} gives each its own row. One atomic test was planned per metric, giving twenty tests per profile, and every verdict was tied to a URL, an excerpt, and a capture hash. The verdicts were adjudicated by a single rater, an AI assistant working from the captured evidence under an author's direction; they have not yet been re-rated independently. This section describes what an agent could observe and makes no finding on any merchant's compliance with any law or rule.

\begin{figure}
\centering
\includegraphics[width=0.86\linewidth,keepaspectratio]{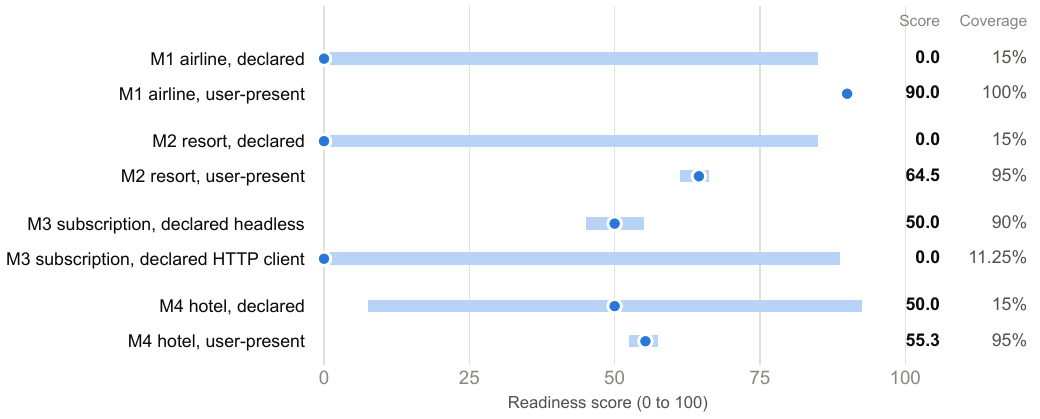}
\caption{Observed readiness score (dot) and missingness bounds (bar) for the eight assessed profiles. The bar spans the score the profile would receive if every unobserved test failed and if every unobserved test passed. Two profiles with the same observed score of 50.0 (M3 declared headless and M4 declared) carry bounds of very different width.}
\label{fig:bounds}
\end{figure}

\textbf{Results.} Table~\ref{tab:pilot} and Figure~\ref{fig:bounds} give the results.

\begin{table}[h]
\centering
\caption{Full assessments, 29 September 2026. One mandate and one observation per merchant; verdicts adjudicated by a single rater. ``Declared'' covers both declared profiles where they met the same barrier at every page (M1, M2, and M4); M3 is reported per declared profile. ``Determinate'' counts the mandate criteria, out of seven, in the DETERMINATE state; ``none'' means that no criterion was observable. The last column applies the publication rule of Section~\ref{sec:scoring}.}
\label{tab:pilot}
\begin{tabular}{>{\raggedright\arraybackslash}p{5.2cm}rr>{\raggedright\arraybackslash}p{2.6cm}c>{\raggedright\arraybackslash}p{2.2cm}}
\toprule
\textbf{Merchant and profile} & \textbf{Score} & \textbf{Coverage} & \textbf{Bounds} & \textbf{Determinate} & \textbf{Rule} \\
\midrule
M1 airline, declared & 0.0 & 15\% & 0 to 85 & none & Not met \\
M1 airline, user-present & 90.0 & 100\% & 90 to 90 & 6 of 7 & Met \\
M2 resort, declared & 0.0 & 15\% & 0 to 85 & none & Not met \\
M2 resort, user-present & 64.5 & 95\% & 61.25 to 66.25 & 6 of 7 & Not met \\
M3 subscription, declared headless & 50.0 & 90\% & 45 to 55 & 3 of 7 & Not met \\
M3 subscription, declared HTTP client & 0.0 & 11.25\% & 0 to 88.75 & none & Not met \\
M4 hotel, declared & 50.0 & 15\% & 7.5 to 92.5 & none & Not met \\
M4 hotel, user-present & 55.3 & 95\% & 52.5 to 57.5 & 5 of 7 & Not met \\
\bottomrule
\end{tabular}
\end{table}

\emph{Get in.} Three of the four merchants refused the declared profiles at the offer pages. Two answered every request on the day, including the request for \texttt{robots.txt}, with a challenge page from a bot-management service. The third served one discovery page and refused its home, offer, and policy pages. The fourth served every public page to the declared headless browser and an empty shell to the plain HTTP client. The user-present browser read all three sites that had refused the declared profiles.

\emph{Decide.} No task reached joint decision sufficiency: each left at least one criterion with money attached undetermined. At M1, the displayed fares were stated to include taxes and carrier charges, and the fare and bundle prices gave a finite total inside the budget, so the total was determinate. The refund was not: one policy page said that a booking canceled within 24 hours is canceled without penalty, and another said that a component of the fare is not refunded. A second rater could reasonably read that pair as determinate and unmet, or as determinate and met if the 24-hour statement governs, and it is the kind of case a two-rater study is meant to settle. At M2, the nightly rate included a stated mandatory charge and no tax basis, which left the total without a stated upper bound. The cancellation rule, by contrast, was stated exactly, and it made the offer unsuitable for a mandate that required free cancellation until the day before arrival. M2 is therefore the one task of the four with a decisive rejection: the agent could not confirm the total, and it did not need to. At M3, the recurring price was determinate, while the hardware price, shipping, and tax were not stated on any page before an order flow that the site's \texttt{robots.txt} disallows, the refund statement covered the hardware alone, and the governing terms were not reachable from the offer page. At M4, a statement of a mandatory nightly fee and its tax treatment appeared for the site's default dates and was absent for the scenario dates, and the cancellation terms were deferred to the reservation confirmation.

\emph{Prove.} Every criterion that was determinate rested, as observed, on evidence of class SELF: the agent's own record of what it saw. No merchant-signed offer was observed before commitment at any of the four sites. The assessment stops before checkout, so it cannot see what a completed purchase would add (a signed checkout session, a confirmation message, the processor's record), and the finding is confined to the evidence available when the agent must decide.

\emph{Information reliability.} The comparisons available in a single visit were few (one to four per merchant), and they turned up conflicts at three of the four merchants: a headline starting price below every price offered for the dates, a fee statement present in one date context and missing in another, and two policy pages that agreed on a cancellation window and differed on what is returned. Structured data was unreliable where it existed: two sites used type names that are not schema.org types, and one emitted unrendered template text as a product name.

\emph{Effort.} Of the effort measures of Section~\ref{sec:sufficiency}, the pilot recorded the pages captured per task: six at M1, three at M2, and eight at M4 in the user-present sessions, and seven at M3 under the declared headless browser. Machine actions, elapsed time, and retries were not recorded.

\emph{Bands.} Under the illustrative rule, every profile that read the pages projects to ELEVATED. That follows from the rule and adds no information: each task left a total or a refund term indeterminate, which is the rule's condition for that band. The profiles that saw nothing fall under CRITICAL by the rule's first line; that band reflects access alone and is not reported as a projection.

\subsection{An access census of 555 sites}
\label{sec:census}

Three of four merchants refused a declared agent. The second part of the pilot asks how far that finding extends.

\textbf{Sample.} We found no published client roster for either of two large payment processors, so the sample was built from public statements. It contains every merchant that either processor has named as a client in its own public materials, as far as we could establish on 29 September 2026, and whose commercial domain could be identified with confidence. After two domains that did not resolve were removed and clients that share a host were merged, the sample held 555 sites. Sector labels were assigned from the vertical recorded for each client. The two client lists differ in sector mix, and the results are reported for the pooled sample and by sector, not by processor.

\textbf{Procedure.} On 30 September 2026, between 11:38 and 12:07 UTC, each site received about five requests under the rules of Section~\ref{sec:collection}: \texttt{robots.txt}, the home page, \texttt{/.well-known/ucp}, and \texttt{/llms.txt} from the declared HTTP client, and the home page from the declared headless browser. All requests came from one network location in Ontario, Canada. The outcome for the home page was classed as \emph{readable}; \emph{challenged or refused} (a challenge page, a block page, a holding page, or an HTTP error status in place of the home page); \emph{empty shell} (a successful response carrying fewer than 200 characters of text); \emph{no response} (\texttt{robots.txt} or the page could not be retrieved and no HTTP status was received: a timeout, a reset, or a connection or certificate error); or \emph{disallowed} by \texttt{robots.txt}. The classification was automatic and was then audited by one rater, who read the title and opening text of every page classed readable, inspected every empty shell, and viewed screenshots of 48 randomly drawn readable pages and 24 challenged ones. The audit changed the class of 14 of the 1,110 home-page observations, at ten sites: a waiting-room page and a CAPTCHA page served with HTTP 200 had been classed readable, for example, and four short but genuine home pages had been classed as empty shells. A later check of every challenge verdict that had not come with a 403 status changed one more: a location-selection page had been classed as a challenge because the name of a bot-management service appeared in its response headers, and it is an empty shell. The corrections are recorded with their reasons.

\textbf{Results.} Table~\ref{tab:census} gives the outcomes.

\begin{table}[h]
\centering
\caption{Home-page outcomes for 555 sites, 30 September 2026, by declared profile: number of sites (percent).}
\label{tab:census}
\begin{tabular}{>{\raggedright\arraybackslash}p{5.0cm}rr}
\toprule
\textbf{Outcome} & \textbf{HTTP client} & \textbf{Headless browser} \\
\midrule
Readable & 418 (75.3) & 423 (76.2) \\
Challenged or refused & 87 (15.7) & 108 (19.5) \\
Empty shell & 24 (4.3) & 0 (0.0) \\
No response & 22 (4.0) & 20 (3.6) \\
Disallowed by robots.txt & 4 (0.7) & 4 (0.7) \\
\midrule
Readable by neither profile & \multicolumn{2}{c}{105 (18.9)} \\
Challenged or refused by both profiles & \multicolumn{2}{c}{76 (13.7)} \\
\bottomrule
\end{tabular}
\end{table}

\begin{figure}
\centering
\includegraphics[width=0.86\linewidth,keepaspectratio]{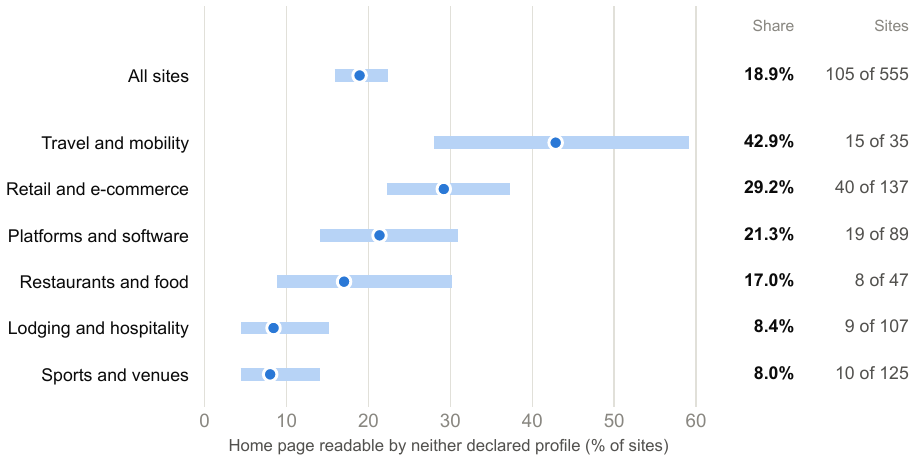}
\caption{Share of sites whose home page neither declared profile could read, with 95\% Wilson intervals, for the whole sample and for the six sectors with at least 20 sites.}
\label{fig:census}
\end{figure}

Most home pages admit a declared agent. About three quarters were readable to each profile, and 391 sites (70.5\%) were readable to both. At 105 sites (18.9\%; 95\% Wilson interval~\cite{wilson1927} 15.9 to 22.4), neither profile could read the home page; 76 sites (13.7\%) challenged or refused both profiles, and 119 sites (21.4\%) challenged or refused at least one of the two. The first figure counts every site that neither profile could read, whatever the cause, and the second only the sites that answered both profiles with a challenge, a block, or an error status. The rate is far from uniform (Figure~\ref{fig:census}). By sector, it ran from 8.0\% of sports and venue sites and 8.4\% of lodging and hospitality sites to 29.2\% of retail sites and 42.9\% of travel and mobility sites, where the interval is wide (15 of 35). The intervals are descriptive, since the sample is a list of named clients and not a random draw. The four merchants of Section~\ref{sec:deepdives} are large brands in lodging, air travel, and subscriptions; their rate of three in four overstates what a declared agent meets across the broader sample.

Where a site challenged or refused a profile, a bot-management service could be identified from the response at 91 of the 119 sites, and one service accounted for 61 of them. The two profiles were not treated alike: 25 sites served the home page to the plain HTTP client and challenged or refused the browser, 10 did the reverse, and 20 returned an empty shell to the HTTP client and a full page to the browser. The request for \texttt{robots.txt} itself met a 403 or 429 status at 63 sites (11.4\%). RFC~9309 treats such a file as unavailable, which permits a crawler to proceed~\cite{rfc9309}; the site has then withheld the rules it expects crawlers to follow.

The protocol files tell a second story. A \texttt{robots.txt} file was readable at 428 sites (77.1\%), close to the 78\% found in a scan of the 200,000 most-visited domains~\cite{cloudflare2026readiness}; of these, 91 (21.3\%) name at least one AI crawler or agent, and 44 (10.3\%) disallow at least one of them entirely. Of the 456 sites where the path could be observed, 108 (23.7\%) published an \texttt{llms.txt} file. Of the 429 sites where that path could be observed, 28 (6.5\%) published a UCP business profile; 24 of the 28 are retail sites, and 21 reference storefronts on a single hosted commerce platform. These files are weak signs of readiness. Forty-seven sites (8.5\%) answered at least one protocol path with HTTP 200 and something other than the protocol document, in all but one case an ordinary web page; one \texttt{llms.txt} file contained nothing but placeholder text; and 12 sites published an \texttt{llms.txt} file while neither declared profile could read their home page.

The census is one observation per site. How stable such observations are was not measured; a repeat of the census on a later day would show it, and the benchmark's I3 and I4 rates exist to measure that stability.

\subsection{A declared assessment of 330 merchants}
\label{sec:fullrun}

\textbf{Sample and procedure.} The sample held 330 merchants with an online store, nearly all retailers, selling in the United States and Canada. They came from three lists: two supplied by a commercial partner of the authors' company that operates a consumer shopping agent, its catalog of stores on one hosted commerce platform (227) and its partner merchants (13), and one of large North American retailers (90). Each merchant was assessed under both declared profiles from one offer page per merchant and profile, chosen by the scanner from the home page, the category links, and the sitemap, together with the home page and the policy pages linked from them, against one retail mandate: quantity one, delivery to ZIP code 10001, a budget of the displayed price plus 25\%, delivery within 10 days, and returns within 14 days. Access and discovery (pillar A) were observed on 3 October 2026 and pillars B to E on 7 October, from pages fetched that day under the rules of Section~\ref{sec:collection}. Each metric was split into named checks, 38 per profile and 25,080 in all, and every status was assigned by deterministic rules (rubric version 0.1); no language model read a page. A check is UNKNOWN when its page could not be read, when no page was found for it, or when the page was read and the rules could not decide. The merchants are not identified, because each result rests on one observation of one offer.

\begin{figure}
\centering
\includegraphics[width=0.86\linewidth,keepaspectratio]{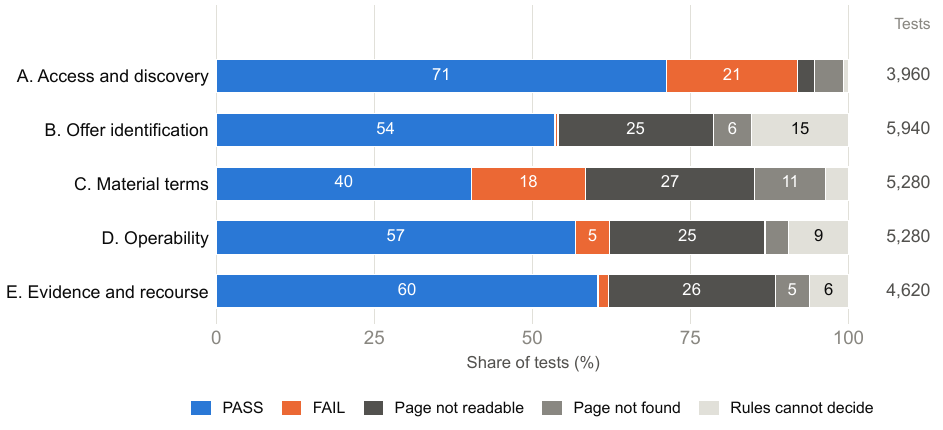}
\caption{Status of the checks of each pillar, both declared profiles pooled, 330 merchants. Pillar A counts a barrier as a failure of access; pillars B to E count it as a page not readable. Numbers are percent of the pillar's checks.}
\label{fig:fullrun-status}
\end{figure}

\begin{figure}
\centering
\includegraphics[width=0.86\linewidth,keepaspectratio]{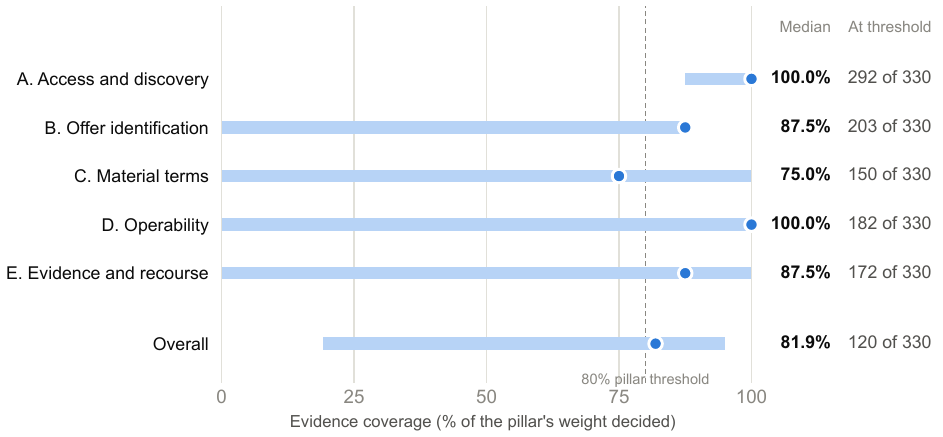}
\caption{Evidence coverage per pillar and overall under the declared HTTP client, 330 merchants: median (dot) and interquartile range (bar). The right column counts profiles at the threshold: 80\% for a pillar, 90\% overall.}
\label{fig:fullrun-coverage}
\end{figure}

\begin{figure}
\centering
\includegraphics[width=0.86\linewidth,keepaspectratio]{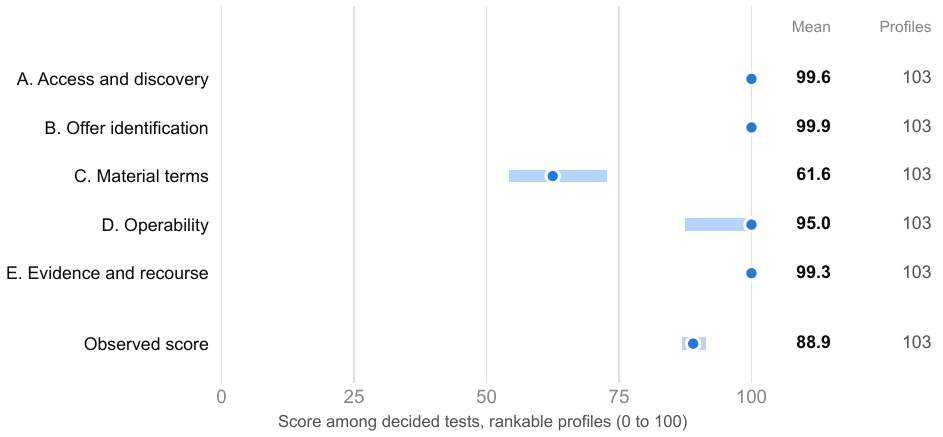}
\caption{Pillar scores of the 103 rankable profiles under the declared HTTP client: median (dot), interquartile range (bar), and mean. Material terms is the one pillar well below the others.}
\label{fig:fullrun-scores}
\end{figure}

\begin{figure}
\centering
\includegraphics[width=0.86\linewidth,keepaspectratio]{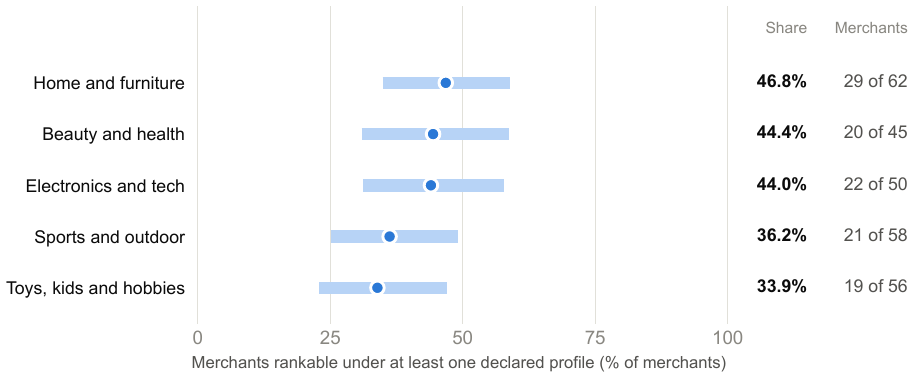}
\caption{Share of merchants rankable under at least one declared profile, with 95\% Wilson intervals, for the five categories with at least 20 merchants. The 59 merchants in smaller categories are not shown; 4 of them were rankable.}
\label{fig:fullrun-category}
\end{figure}

\textbf{Results.} Table~\ref{tab:fullrun} and Figures~\ref{fig:fullrun-status} to~\ref{fig:fullrun-category} give the results. Coverage, not the observed score, decided who could be ranked. The median profile had 82\% of its weight decided, but a rankable profile needs 90\% overall and 80\% in every pillar, and 103 of the 330 profiles of the HTTP client and 102 of the headless browser reached that; 115 merchants were rankable under at least one profile and 90 under both. The profiles that fell short mostly fell short in pillars B to E together (Figure~\ref{fig:fullrun-status}): of the 8,676 unknown checks in those pillars, 62\% were on a page the profile could not read, 16\% had no page, because the scanner found no offer page or no policy page, and 22\% were on a page the rules could not decide. Material terms had the lowest coverage (a median of 75\% under the HTTP client, with 150 profiles at the pillar threshold; Figure~\ref{fig:fullrun-coverage}), and most of its failures came from one check, the all-in total: under the HTTP client it was determinate at 18 merchants and failed at 178, 137 of them on tax, which was not stated on the offer or policy pages, stated without a rate, or deferred to checkout, and 41 on shipping.

\begin{table}[h]
\centering
\caption{Declared assessment of 330 merchants. Pillar A observed 3 October 2026, pillars B to E on 7 October; rubric version 0.1. Values are medians over profiles unless stated. A profile is rankable at 90\% coverage overall and 80\% in every pillar.}
\label{tab:fullrun}
\begin{tabular}{>{\raggedright\arraybackslash}p{7.0cm}rr}
\toprule
 & \textbf{HTTP client} & \textbf{Headless browser} \\
\midrule
Profiles & 330 & 330 \\
Coverage & 81.9\% & 82.7\% \\
Observed score & 87.5 & 87.0 \\
Rankable profiles & 103 & 102 \\
\quad observed score & 89.0 & 90.2 \\
\quad bounds & 86.0 to 89.6 & 86.3 to 90.6 \\
\quad mean pillar score A / B / C / D / E & 99.6 / 99.9 / 61.6 / 95.0 / 99.3 & 99.8 / 99.8 / 63.7 / 94.6 / 99.6 \\
\midrule
Merchants rankable under at least one profile & \multicolumn{2}{c}{115 of 330} \\
Merchants rankable under both profiles & \multicolumn{2}{c}{90 of 330} \\
\bottomrule
\end{tabular}
\end{table}

Among the rankable profiles, four pillars scored 95 or more on average and material terms scored 62 under the HTTP client and 64 under the headless browser (Figure~\ref{fig:fullrun-scores}). The median rankable profile under the HTTP client scored 89.0, with bounds of 86.0 to 89.6: once coverage is high, the bounds are narrow, and what remains open is concentrated in one pillar. By category (Figure~\ref{fig:fullrun-category}), the share of merchants rankable under at least one profile ran from 34\% of toys, kids, and hobbies merchants to 47\% of home and furniture merchants, with overlapping intervals; the median rankable score was between 88.6 and 90.3 in every category, and 4 of the 59 merchants in smaller categories were rankable.

\textbf{Repeat.} Pillars B to E were measured again on 8 October 2026 from pages fetched that day, with the pillar A observation of 3 October unchanged. Of the 21,120 checks of pillars B to E, 323 (1.5\%) changed status, at 28 merchants: 252 moved from unknown to decided, 67 the other way, and 4 swapped between pass and fail. The access class of the offer page differed between the days at 13 merchants, most often a \texttt{robots.txt} file unreachable on one day and readable on the other. The 104 profiles of the HTTP client that were rankable on the second day include all 103 of the first, each with the same observed score; of the headless-browser profiles, 100 were rankable on both days and 96 kept their score, the largest move being 2.5 points. One merchant entered the set rankable under at least one profile and none left; 90 merchants were rankable under both profiles on each day, two entering and two leaving that set.

\section{Discussion}
\label{sec:discussion}

\emph{Access and sufficiency are separate.} In the four assessments, the user-present browser passed every barrier that had stopped the declared profiles, and none of the three mandates it was run on could be confirmed as met. What remained unresolved once the pages were readable was a total or a refund term. A readiness measure that stops at access, markup, or protocol files would not have seen that.

\emph{Declared agents are often refused, for reasons this study cannot attribute.} At three of the four merchants, pages that refused the declared profiles opened in the user-present browser. In the census, about one home page in five was readable to neither declared profile. The census sent no undeclared request, and the user-present sessions differed from the declared ones in more than the declaration, so neither result shows that declaring was the cause. They do give a baseline. The mechanisms that would let a site admit an agent because of who it is now exist: signed requests~\cite{meunier2026webbotauth}, network-level agent identification~\cite{visa_tap2025}, and bot-management categories for agents acting for a person~\cite{cloudflare2026aitraffic}. The probes used here carried no signature and were registered with no service, and repeating the census with a signed identity would measure how much of the gap verification closes. The question has weight beyond measurement, since the appellate case cited above concerned an agent whose developer chose not to announce it~\cite{amazon_perplexity2026}.

\emph{The gaps were ordinary ones.} A tax basis was not stated beside a nightly rate; a fee statement appeared for one set of dates and not another; two policy pages disagreed about what a cancellation returns; charges were not stated before an order flow. None requires a new protocol to repair. The same gap appears at scale: in the declared assessment of 330 merchants, the all-in total was determinate at 18 merchants under the HTTP client, and tax was the term most often left open. Each leaves an agent with a hard budget or a refund condition unable to proceed, or able to proceed only by assuming. Whether such gaps lead to failed purchases or disputes is not something the pilot can show.

\emph{Protocol files are not readiness.} A quarter of the observable sites publish an \texttt{llms.txt} file, and a small group, most of it on one platform, publishes a UCP profile. Some of the same sites refuse a declared agent, answer protocol paths with web pages, or publish a file of placeholder text. The presence of a file is a diagnostic, which is why the benchmark records it outside the score.

The measurements are meant to be used in three ways: by a merchant, as a repair list ordered by weight; by a payment processor or acquirer, as a distribution across its merchants, with a Merchant Clearing Passport whose outcome fields its own records can fill; and by an agent, as a record it can query during a task to choose where to try first and what to warn the person about. In each use the record carries its observation date, its coverage, and its bounds, and it informs. It does not replace the person's approval, and it must not release a payment.

\section{Limitations and Validation}
\label{sec:limitations}

The full assessments cover four merchants, one mandate each, observed once and adjudicated by a single rater. The rater was not independent of the design: the same AI assistant, under the authors' direction, helped write the rubric, collected the user-present evidence, adjudicated the verdicts, and audited the census. No inter-annotator agreement has been measured, so the determinacy judgments carry one reading of each page, and at least one of them (the refund at M1) is contestable. The rules of Table~\ref{tab:determinacy} were written from these same cases and have not been applied by a second rater to unseen ones; they do not yet say whether an agent may use outside knowledge, such as a statutory tax rate, to bound a total. The results show that the constructs can be applied and what they expose. They are not a benchmark result for any merchant, and the merchants are not ranked.

The scores are coarse. With one test per metric, a pillar moves in steps of 25 points, and a single unknown test can fail the per-pillar coverage rule. The weights are unfitted, and no sensitivity analysis has been run. The band rule is uncalibrated and, in the pilot, circular. Nothing in this paper shows that an indeterminate total or refund term causes disputes, refunds, or losses; that claim requires transaction outcomes, which only a processor or a merchant holds. The evidence classes are reported for the moment of decision only, because the assessment stops before checkout.

The census measures the home page only, on one day, from one network location, for two declared profiles that carry no cryptographic signature and are registered with no bot-management service. It says nothing about search, offer, or booking pages, where defenses often differ, or about what an endpoint built for agents returns. It cannot separate a refusal aimed at a declared agent from one that any automated client, or any visitor from that network, would meet, because the study deliberately sent no undeclared control request; sites that restrict access by geography count as refusals. A site that refuses an unknown declared bot may admit a verified one. The no-response class mixes timeouts, resets, and certificate errors, some of which would affect any visitor. The 200-character threshold for an empty shell is a convenience, and the audit corrected four pages it misclassed. The classification was audited by one rater, and the sector labels were assigned after the fact from the vertical recorded for each client. The sample consists of merchants that two processors have publicly named, which favors large brands; it is not a random sample of either processor's merchants, and a client relationship does not establish which processor handles a site's web checkout.

The declared assessment of 330 merchants is one observation of one offer page per merchant and profile, chosen by a scanner that recognizes retail offer pages, so a merchant whose offer page it did not find is unknown on pillars B to E, and the assessment says nothing about other offers of the same merchant. Its statuses come from deterministic rules written for retail pages, under rubric version 0.1, whose choices stand for this version and are stated here with their size. A tax that a page says applies, without an amount or a rate, leaves the all-in total open, and no published rate is used: the total check passed at 18 merchants under each profile and failed on tax at 137 under the HTTP client and 134 under the headless browser. One absent tax statement fails both the tax-basis check and the total check, since they ask different questions, whether the basis is disclosed and whether the total can be decided. The rubric does not settle whether a price with no stated unit is per item, or whether a regular price shown beside a sale price states a condition, so those two checks stayed unknown at 203 and 185 profiles of the HTTP client and 198 and 194 of the headless browser. The scan ran from Ontario, Canada, and terms were judged as served: under the HTTP client, 30 merchants served a price in Canadian dollars or pounds, 12 of them rankable, and 128 material-terms passes rest on such pages (10 merchants, 4 rankable, and 41 passes under the headless browser), so a result can describe a market other than the mandate's. Six checks have no observable failure under the current rules (the item's name, the charging basis, price conditions, availability for the selected configuration, the seller's name, and the information required for a return): they pass or remain unknown, so when decided they can only raise the observed score. They carry 13.1\% of the weight, and without them the mean observed score of the rankable profiles would be 87.7 instead of 89.0 under the HTTP client and 88.2 instead of 89.5 under the headless browser.

An audit of 192 material-terms verdicts, drawn by stratum from the 7 and 8 October runs and rated from the stored pages by an AI model acting as one rater, agreed with the rules on 120 (62\%): on 85\% of the passes and 57\% of the failures, falling to 28\% where the rules reported conflicting terms, most often a tiered or scoped clause of one policy read as a second policy. The patterns mostly under-score, failing terms that the page states, and at seven merchants an absence was judged on a page that holds no policy text. Correcting the misreads the audit identified would move the mean material-terms score of the rankable profiles by one to three points and no merchant's rank. The corrections are not applied in this version, no person has reviewed the disagreements, and the audit's rates describe its strata, not the run.

The user-present profile is one browser operated by one assistant with one person. It shows that a user-present agent was not stopped where declared agents were; it does not characterize consumer agents in general, whose treatment by bot-management systems varies and is changing~\cite{wang2026fpagent,cloudflare2026aitraffic}. The RAILS constructs are applied by extension: the admissibility classes were defined for evidence about an agent's performance of an obligation, and applying SIGN to a merchant's signature over its own offer, or treating a processor record as REC, is an interpretation that a deployment would have to adopt explicitly. Anonymizing the merchants limits independent replication of the four assessments; the capture hashes, excerpts, and records are retained.

Validation has to begin with the measurement itself~\cite{bean2025construct}. Two raters will independently annotate a stratified subset of tests, including the unknowns and every blocking flag, with agreement reported per metric by a chance-corrected statistic~\cite{cohen1960kappa}, and the dictionary will be frozen before any comparison between merchants is reported. The declared profiles read pages and three protocol paths; a merchant or its platform may also publish an endpoint built for agents, such as a UCP manifest or an MCP server, whose answers can differ from the pages, and a later version assesses what that endpoint returns against the same mandate and reports it as a route beside the three profiles. A first cohort of 20 to 30 merchants in one vertical, with about 20 offers each and three scheduled observations, would yield the repeat rates and a first distribution of joint decision sufficiency. The question that decides whether the measurements matter is whether determinacy, contradiction rates, and joint sufficiency improve held-out prediction of agent task errors and verification effort beyond simple baselines such as the commerce platform, the vertical, and structured-data completeness. Calibrating the clearing band needs transaction outcomes with denominators and maturation windows; until such data exist, the band remains an illustration, and the Passport's outcome fields remain empty.

\section{Conclusion}

Agent benchmarks fix the tasks and vary the agent. This paper fixes the task and varies the merchant, and it asks three things of a merchant's website: whether an agent can get in, whether the published terms decide the purchase, and whether anything would later prove what was agreed. The pilot suggests that access and decision sufficiency are separate properties. Most of the 555 sites examined let a declared agent read the home page, and about one in five did not; the four merchants examined in depth all left a total or a refund term undetermined once their pages were read; of 330 merchants assessed by rule, about one in three published enough for a ranked score, and among those the all-in total was the term that most often stayed open; and before commitment, the only record of the displayed terms was the agent's own. The measurements needed to see this are simple to state: a mandate, a state for each criterion, a score that carries its coverage and bounds, and a class for each piece of evidence. Whether they predict failed purchases and disputes is the question that outcome data will answer.

\section*{Ethics statement}

The declared profiles identified themselves, followed \texttt{robots.txt}, waited a stated crawl delay up to a cap of 120 seconds, sent about five requests to each site at about one request per second, and requested no cart, checkout, booking, or login URL. One site asked for a longer delay than the cap; it received a single home-page request from each profile after a 120-second wait and no further requests. The profiles did not attempt to pass any barrier: no challenge was solved, no identity was forged, and no request was retried in another guise. Each planned URL was requested once per profile, including after an earlier URL at the same site had been refused. In the declared assessment of 330 merchants, each profile sent about eleven requests per site over the two observation days, at most eighteen: \texttt{robots.txt}, the home page, the sitemap, one offer page, the policy pages, and the protocol paths.

The user-present sessions ran in the person's ordinary browser, operated by an AI browser assistant that the browser's user agent does not announce, with the person present and approving each step, on three sites that had refused the declared profiles. They were read-only views of public pages that the person could have opened by hand: no account was created, no personal data was submitted, and no order, hold, or booking was placed. We regard these sessions as the person's own browsing with assistance, and we report the configuration plainly because a site may regard it differently. Merchants and processors are not identified, because the results rest on single observations. The study observed public web pages, collected no personal data, and placed no order, and no institutional review was sought. A merchant that does not wish its site observed in later runs can write to the authors, and the declared profiles honor a \texttt{robots.txt} rule that names them. The design follows the principles of the Menlo Report~\cite{menlo2012}.

\section*{Data and code availability}

The rubric (Appendix~\ref{app:metrics}), the census records with site identities removed, the capture hashes of the full assessments, and the aggregate results of the declared assessment of 330 merchants are released as a dataset at https://doi.org/10.5281/zenodo.23325184. The scanner is available from the authors on request.

\section*{Competing interests}

The authors are affiliated with Evolutionairy AI, which develops RAILS and has a commercial interest in merchant assessment. No merchant or payment processor commissioned, funded, or reviewed this work. Two of the three merchant lists of Section~\ref{sec:fullrun} were supplied by a commercial partner of Evolutionairy AI that operates a consumer shopping agent; the partner did not fund, review, or direct the work, and it is not identified because the merchants are not.

\section*{Use of AI tools}

The concepts, the benchmark design, and the entire content are the authors'. The scanning and analysis code was developed with an AI assistant (Claude, Anthropic) working under the authors' direction. The same assistant supported the literature search, adjudicated the pilot verdicts from the captured evidence, and audited the census classification. The user-present observations were collected by an AI browser assistant, with an author present to approve each step. In the declared assessment of 330 merchants no language model read a page or assigned a status; a second AI model (Claude Opus 5.5, Anthropic) rated the 192 verdicts of the material-terms audit from the stored pages, as one rater, to check the rules. The authors take responsibility for the entire content.

\appendix

\section{Metric dictionary}
\label{app:metrics}

\begin{table}[h]
\centering
\caption{Pass conditions of the twenty readiness metrics (rubric version 0.1).}
\label{tab:dictionary}
\small
\begin{tabular}{p{0.6cm}>{\raggedright\arraybackslash}p{3.3cm}>{\raggedright\arraybackslash}p{11.4cm}}
\toprule
\textbf{ID} & \textbf{Measurement} & \textbf{Observable pass condition} \\
\midrule
A1 & Readable route & An eligible offer or policy URL returns usable content through the permitted assessment route; HTTP 200 alone is insufficient. \\
A2 & Offer discovery & A seeded eligible offer can be reached from the home page, category links, sitemap, or declared catalog route within the fixed discovery budget (three sitemap files and three candidate pages). \\
A3 & Offer content exposure & The selected route exposes the offer name, selected configuration, and displayed price as text or typed data, and not only as an image or an unresolved shell. \\
A4 & Navigation resolves & The next offer or policy destination is identifiable from a link or accessible control name and resolves to the intended destination. \\
\midrule
B1 & Offer identity & The record identifies the merchant or location and an unambiguous item or service, with a stable identifier where appropriate. \\
B2 & Configuration specificity & The record binds the price to the actual variant, room and rate plan, dates, party size, quantity, modifiers, or subscription tier used in the scenario. \\
B3 & Price interpretation & The record gives amount, currency, and charging basis, with membership, promotional, per-night, or recurring conditions explicit. \\
B4 & Availability & The record exposes stock, bookability, or service eligibility for the selected configuration and context. A clear unavailable result also passes. \\
\midrule
C1 & Cost components disclosed & For the scenario, the site states the base charge and the applicability of mandatory tax, shipping, service, resort, or other fees, including inputs still required. \\
C2 & Total cost determinable & The permitted route yields a numerical total or a valid finite bound that supports the scenario's budget decision. Unknown taxes or fees cannot silently become zero. \\
C3 & Delivery or service determinacy & The applicable destination and delivery window, or the service dates, location, and operating constraints, are explicit. Generic speed claims do not establish arrival by a required date. \\
C4 & Return or cancellation determinacy & Eligibility, deadline, fees, exceptions, and refund method are explicit enough to evaluate the scenario. Clear non-refundable terms pass. \\
\midrule
D1 & Action target identifiable & The purchase or booking entry control and its target are identifiable, named, and associated with the intended offer. Presence is not successful execution. \\
D2 & Selection state visible & The current variant, date, quantity, or modifiers, and any disabled state, are exposed in the rendered page or declared interface. \\
D3 & Prerequisites visible & Required inputs and restrictions, such as destination, account, age check, or membership, are disclosed before the point at which they prevent the intended step. \\
D4 & Commerce handoff resolves & The visible ordering or booking handoff resolves to an identifiable operator and preserves an observable association with the merchant or location. \\
\midrule
E1 & Counterparty disclosed & The site identifies the seller or service operator and contact details sufficient to attribute the offer. \\
E2 & Policy applies to this offer & The relevant policy is linked, or its scope explicitly covers the sampled offer, location, rate plan, and jurisdiction; a generic policy page is insufficient when exceptions apply. \\
E3 & Offer referenceable & The offer has a URL or declared reference that resolves to the identified offer and configuration, or includes explicit reconstruction parameters. A hash generated by the crawler earns no merchant points. \\
E4 & Recourse instructions & The applicable return, cancellation, or dispute instructions identify a usable channel and the information required. This establishes instructions, not response quality. \\
\bottomrule
\end{tabular}
\end{table}

\section{Observation record}
\label{app:record}

Each observation is stored with the following fields: \texttt{merchant\_id}, \texttt{location\_id}, \texttt{origin}, \texttt{final\_url}, \texttt{access\_class}, \texttt{wall\_vendor}, \texttt{commerce\_operator}, \texttt{profile\_id}, \texttt{login\_state}, \texttt{scenario\_id}, \texttt{offer\_id}, \texttt{variant\_or\_rate\_plan}, \texttt{metric\_id}, \texttt{test\_id}, \texttt{planned\_weight}, \texttt{status}, \texttt{reason\_code}, \texttt{observed\_at\_utc}, \texttt{locale}, \texttt{destination}, \texttt{vantage}, \texttt{currency}, \texttt{source\_pointer}, \texttt{source\_excerpt}, \texttt{capture\_hash}, \texttt{capture\_scope}, \texttt{extractor\_version}, \texttt{rubric\_version}, \texttt{review\_state}, and \texttt{signal\_origin}.

The field \texttt{profile\_id} takes the values \texttt{declared-http}, \texttt{declared-headless}, and \texttt{user-present-browser}. The field \texttt{access\_class} takes the values \texttt{OK}, \texttt{BOT\_WALL} (with \texttt{wall\_vendor}), \texttt{EMPTY\_SHELL}, \texttt{HTTP\_}$n$, \texttt{ROBOTS\_DISALLOWED}, \texttt{ROBOTS\_UNREACHABLE}, \texttt{NO\_RESPONSE}, \texttt{NOT\_FETCHED\_CRAWL\_DELAY}, \texttt{NOT\_FETCHED\_BY\_POLICY}, and \texttt{NOT\_ASSESSED}. The field \texttt{capture\_scope} states whether a hash covers the full response or only an excerpt. The field \texttt{vantage} names the network location of the observation, and \texttt{signal\_origin} takes the values \texttt{platform-default}, \texttt{merchant-configured}, and \texttt{unknown}; neither was recorded in the pilot. The field \texttt{login\_state} was ``none'' for every observation reported here, since no profile logs in, and the scanner version named in Section~\ref{sec:benchmark} is carried as \texttt{extractor\_version}.

\section{Illustrative band projection}
\label{app:bands}

Table~\ref{tab:bands} gives the illustrative rule referred to in Section~\ref{sec:passport}. It is not calibrated, reports no probability, and must not change how any payment is treated until it has been validated against outcomes.

\begin{table}[h]
\centering
\caption{Illustrative projection of the RAILS policy band from the criterion states. The rule is not calibrated and reports no probability.}
\label{tab:bands}
\begin{tabular}{p{2.2cm}>{\raggedright\arraybackslash}p{13.0cm}}
\toprule
\textbf{Band} & \textbf{Illustrative condition} \\
\midrule
CRITICAL & No mandate criterion is observable under the profile. Clearing would need a human arbiter and delayed finality. \\
ELEVATED & Criteria are observable, but a total, cancellation, or return term is indeterminate. Clearing would need a multi-verifier mesh and an escrow delay. \\
MODERATE & Every criterion is determinate, but the financially material criteria rest only on SELF or SIGN evidence. A short appeal window applies. \\
LOW & Every criterion is determinate, and the financially material criteria are supported at or above the floor, for example by REC and WIT. Settlement can be instant. \\
\bottomrule
\end{tabular}
\end{table}

\section{A worked mandate}
\label{app:mandate}

Table~\ref{tab:worked} gives the criterion table for M1 under the user-present profile. The mandate was: book a round trip for one adult between two named airports on two fixed dates, in the standard cabin with one carry-on bag, for an all-in total within a fixed budget, with a full refund to the card if canceled within 24 hours of booking.

\begin{table}[h]
\centering
\caption{Criterion states and evidence classes for M1, user-present profile. ``Today'' is the best class obtainable before payment as the site stands; ``With added evidence'' assumes a merchant-signed offer, an independent witness capture, and the processor's record.}
\label{tab:worked}
\begin{tabular}{>{\raggedright\arraybackslash}p{3.6cm}>{\raggedright\arraybackslash}p{2.6cm}p{1.2cm}>{\raggedright\arraybackslash}p{2.0cm}>{\raggedright\arraybackslash}p{5.4cm}}
\toprule
\textbf{Criterion} & \textbf{State} & \textbf{Today} & \textbf{With added evidence} & \textbf{Basis} \\
\midrule
Flights and dates & DETERMINATE & SELF & WIT & Flight numbers, dates, and times shown \\
Passenger and fare class & DETERMINATE & SELF & WIT & One adult, standard cabin \\
All-in total within budget & DETERMINATE & SELF & REC & Fares stated to include taxes and carrier charges; with the bundles that include a carry-on, the total is inside the budget \\
Seats available & DETERMINATE & SELF & SIGN & Remaining seats at the fare shown \\
Full refund within 24 hours & INDETERMINATE & SELF & WIT & ``Without penalty'' on one page; a component of the fare ``not refunded'' on another \\
Operator of record & DETERMINATE & SELF & WIT & Operating carrier named in the site's terms; a processor record would name the merchant of record, not the operator \\
Recourse channel & DETERMINATE & SELF & WIT & Reservations phone line and online cancellation \\
\bottomrule
\end{tabular}
\end{table}

Six of seven criteria are determinate, the task does not have joint decision sufficiency, and the one indeterminate criterion is the refund. Stating whether a cancellation within 24 hours returns the full displayed price would make the seventh criterion determinate.


\begin{thebibliography}{99}\raggedright

\bibitem{openai_agent2025} OpenAI (2025). \emph{Introducing ChatGPT agent: bridging research and action}. 17 July 2025. \url{https://openai.com/index/introducing-chatgpt-agent/}.

\bibitem{muse2026} Meta (2026). \emph{Introducing Muse: The World's First Personal AI Agent Built for Everyone}. Meta Newsroom, 8 September 2026. \url{https://about.fb.com/news/2026/09/introducing-muse-personal-ai-agent/}.

\bibitem{openai_discovery2026} OpenAI (2026). \emph{Powering Product Discovery in ChatGPT}. 24 March 2026. \url{https://openai.com/index/powering-product-discovery-in-chatgpt/}.

\bibitem{zhou2023webarena} Zhou, S., Xu, F. F., Zhu, H., et al. (2023). \emph{WebArena: A Realistic Web Environment for Building Autonomous Agents}. arXiv:2307.13854. \url{https://arxiv.org/abs/2307.13854}.

\bibitem{deng2023mind2web} Deng, X., Gu, Y., Zheng, B., et al. (2023). \emph{Mind2Web: Towards a Generalist Agent for the Web}. arXiv:2306.06070. \url{https://arxiv.org/abs/2306.06070}.

\bibitem{he2024webvoyager} He, H., Yao, W., Ma, K., et al. (2024). \emph{WebVoyager: Building an End-to-End Web Agent with Large Multimodal Models}. arXiv:2401.13919. \url{https://arxiv.org/abs/2401.13919}.

\bibitem{xue2025illusion} Xue, T., Qi, W., Shi, T., et al. (2025). \emph{An Illusion of Progress? Assessing the Current State of Web Agents}. arXiv:2504.01382. \url{https://arxiv.org/abs/2504.01382}.

\bibitem{zhang2026clawbench} Zhang, Y., Wang, Y., Zhu, Y., et al. (2026). \emph{ClawBench: Can AI Agents Complete Everyday Online Tasks?} arXiv:2604.08523. \url{https://arxiv.org/abs/2604.08523}.

\bibitem{wang2025shoppingbench} Wang, J., Xiao, K., Sun, Q., Zhao, H., Luo, T., Zhang, J. D., \& Zeng, X. (2025). \emph{ShoppingBench: A Real-World Intent-Grounded Shopping Benchmark for LLM-based Agents}. arXiv:2508.04266. \url{https://arxiv.org/abs/2508.04266}.

\bibitem{lyu2025deepshop} Lyu, Y., Zhang, X., Yan, L., de Rijke, M., Ren, Z., \& Chen, X. (2025). \emph{DeepShop: A Benchmark for Deep Research Shopping Agents}. arXiv:2506.02839. \url{https://arxiv.org/abs/2506.02839}.

\bibitem{peeters2025webmall} Peeters, R., Steiner, A., Schwarz, L., Caspary, J. Y., \& Bizer, C. (2025). \emph{WebMall: A Multi-Shop Benchmark for Evaluating Web Agents}. arXiv:2508.13024. \url{https://arxiv.org/abs/2508.13024}.

\bibitem{tou2025shoppingcomp} Tou, H., Zeng, Y., Li, Y., et al. (2025). \emph{ShoppingComp: Are LLMs Really Ready for Your Shopping Cart?} arXiv:2511.22978. \url{https://arxiv.org/abs/2511.22978}.

\bibitem{allouah2025aces} Allouah, A., Besbes, O., Figueroa, J. D., Kanoria, Y., \& Kumar, A. (2025). \emph{What Is Your AI Agent Buying? Evaluation, Biases, Model Dependence, \& Emerging Implications for Agentic E-Commerce}. arXiv:2508.02630. \url{https://arxiv.org/abs/2508.02630}.

\bibitem{longpre2024consent} Longpre, S., Mahari, R., Lee, A., et al. (2024). \emph{Consent in Crisis: The Rapid Decline of the AI Data Commons}. arXiv:2407.14933. \url{https://arxiv.org/abs/2407.14933}.

\bibitem{liu2024somesite} Liu, E., Luo, E., Shan, S., Voelker, G. M., Zhao, B. Y., \& Savage, S. (2025). Somesite I Used To Crawl: Awareness, Agency and Efficacy in Protecting Content Creators From AI Crawlers. \emph{Proceedings of the 2025 ACM Internet Measurement Conference}, 78--99. \url{https://doi.org/10.1145/3730567.3732913}.

\bibitem{hoffmann2026aitxt} Hoffmann, F., Goergens, T., Khosla, S., \& Bajpai, V. (2026). From robots.txt to ai.txt: Mapping the Evolution of Web Permissions in the Age of AI. \emph{ACM SIGCOMM Computer Communication Review}, 56(2), 29--41. \url{https://doi.org/10.1145/3831956.3831960}.

\bibitem{cloudflare2026readiness} Cloudflare (2026). \emph{Introducing the Agent Readiness score. Check to see if your site is agent-ready}. The Cloudflare Blog, 17 April 2026. \url{https://blog.cloudflare.com/agent-readiness/}.

\bibitem{shopify2026readiness} Shopify (2026). \emph{Product Page Audit for Agentic Commerce}. Accessed 30 September 2026. \url{https://www.shopify.com/agentic-readiness}.

\bibitem{serviloq2026} Serviloq (2026). \emph{Agentic commerce readiness, explained}. Accessed 30 September 2026. \url{https://serviloq.com/learn/readiness}.

\bibitem{ottom2026} Ottom (2026). \emph{How Ottom measures category benchmarks}. Accessed 30 September 2026. \url{https://ottom.io/benchmarks/methodology}. Agent policy: \url{https://ottom.io/agent}.

\bibitem{visa_rules2026} Visa (2026). \emph{Visa Core Rules and Visa Product and Service Rules}. 18 April 2026. \url{https://usa.visa.com/dam/VCOM/download/about-visa/visa-rules-public.pdf}.

\bibitem{mastercard_vi2026} Mastercard (2026). \emph{Verifiable Intent: Open specification for cryptographic agent authorization in commerce}. Draft v0.1. \url{https://verifiableintent.dev/}.

\bibitem{parikh2025ap2} Parikh, S., \& Surapaneni, R. (2025). \emph{Powering AI commerce with the new Agent Payments Protocol (AP2)}. Google Cloud Blog, 16 September 2025. \url{https://cloud.google.com/blog/products/ai-machine-learning/announcing-agents-to-payments-ap2-protocol}.

\bibitem{louck2026signing} Louck, Y., Dvir, A., \& Stulman, A. (2026). \emph{Signing the Transaction but Not the Decision: Whisper Attacks and a Binding Defense for AP2}. arXiv:2609.11757. \url{https://arxiv.org/abs/2609.11757}.

\bibitem{zhang2026sokpayments} Zhang, Y., Xiang, Y., Lei, Y., et al. (2026). \emph{SoK: Blockchain Agent-to-Agent Payments}. arXiv:2604.03733. \url{https://arxiv.org/abs/2604.03733}.

\bibitem{rails2026} de Valois-Franklin, A., \& Bogdan, A. (2026). \emph{RAILS: Verification-Native Clearing for Agentic Commerce}. arXiv:2606.08790. \url{https://arxiv.org/abs/2606.08790}.

\bibitem{yao2022webshop} Yao, S., Chen, H., Yang, J., \& Narasimhan, K. (2022). \emph{WebShop: Towards Scalable Real-World Web Interaction with Grounded Language Agents}. arXiv:2207.01206. \url{https://arxiv.org/abs/2207.01206}.

\bibitem{garg2025real} Garg, D., VanWeelden, S., Caples, D., et al. (2025). \emph{REAL: Benchmarking Autonomous Agents on Deterministic Simulations of Real Websites}. arXiv:2504.11543. \url{https://arxiv.org/abs/2504.11543}.

\bibitem{bansal2025magentic} Bansal, G., Hua, W., Huang, Z., et al. (2025). \emph{Magentic Marketplace: An Open-Source Environment for Studying Agentic Markets}. arXiv:2510.25779. \url{https://arxiv.org/abs/2510.25779}.

\bibitem{ersoy2025darkpatterns} Ersoy, D., Lee, B., Shreekumar, A., Arunasalam, A., Ibrahim, M., Bianchi, A., \& Celik, Z. B. (2025). \emph{Investigating the Impact of Dark Patterns on LLM-Based Web Agents}. arXiv:2510.18113. \url{https://arxiv.org/abs/2510.18113}.

\bibitem{cuvin2025decepticon} Cuvin, P., Zhu, H., \& Yang, D. (2025). \emph{DECEPTICON: How Dark Patterns Manipulate Web Agents}. arXiv:2512.22894. \url{https://arxiv.org/abs/2512.22894}.

\bibitem{kim2025scrapers} Kim, T., Bock, K., Luo, C., Liswood, A., Poroslay, C., \& Wenger, E. (2025). Scrapers Selectively Respect robots.txt Directives: Evidence From a Large-Scale Empirical Study. \emph{Proceedings of the 2025 ACM Internet Measurement Conference}, 541--557. \url{https://doi.org/10.1145/3730567.3764471}.

\bibitem{howard2024llmstxt} Howard, J. (2024). \emph{The /llms.txt file, v2}. Published 3 September 2024, modified 10 August 2026. \url{https://llmstxt.org/}.

\bibitem{belson2025radar} Belson, D. (2025). \emph{The 2025 Cloudflare Radar Year in Review: The rise of AI, post-quantum, and record-breaking DDoS attacks}. The Cloudflare Blog, 15 December 2025. \url{https://blog.cloudflare.com/radar-2025-year-in-review/}.

\bibitem{cloudflare2026aitraffic} Cloudflare (2026). \emph{New options to manage AI traffic}. Changelog, 1 July 2026. \url{https://developers.cloudflare.com/changelog/post/2026-07-01-ai-traffic-options/}.

\bibitem{amazon_perplexity2026} \emph{Amazon.com Services, LLC v.\ Perplexity AI, Inc.}, No.~26-1444 (9th Cir.\ Aug.\ 4, 2026). \url{https://cdn.ca9.uscourts.gov/datastore/opinions/2026/08/04/26-1444.pdf}.

\bibitem{wang2026fpagent} Wang, E., Shafiq, Z., \& Vekaria, Y. (2026). \emph{FP-Agent: Fingerprinting AI Browsing Agents}. arXiv:2605.01247. \url{https://arxiv.org/abs/2605.01247}.

\bibitem{ousat2026brokengates} Ousat, B., Turkmen, N., Rampersaud, L., Bailey, D., \& Kharraz, A. (2026). \emph{Broken Gates: Re-evaluating Web Bot Defenses in the Age of LLM Agents}. arXiv:2607.18659. \url{https://arxiv.org/abs/2607.18659}.

\bibitem{meunier2026webbotauth} Meunier, T., \& Major, S. (2026). \emph{HTTP Message Signatures for automated traffic}. Internet-Draft draft-ietf-webbotauth-httpsig-protocol-00, IETF, 1 September 2026. Work in progress. \url{https://datatracker.ietf.org/doc/draft-ietf-webbotauth-httpsig-protocol/}.

\bibitem{marro2025permission} Marro, S., Chan, A., Ren, X., et al. (2025). \emph{Permission Manifests for Web Agents}. arXiv:2601.02371. \url{https://arxiv.org/abs/2601.02371}.

\bibitem{chowdhury2026termstxt} Chowdhury, R. (2026). \emph{terms.txt: A Consent and Compensation Protocol for Agentic Web Access}. arXiv:2609.11152. \url{https://arxiv.org/abs/2609.11152}.

\bibitem{lu2025buildweb} L\`u, X. H., Kamath, G., Mosbach, M., \& Reddy, S. (2025). \emph{Build the web for agents, not agents for the web}. arXiv:2506.10953. \url{https://arxiv.org/abs/2506.10953}.

\bibitem{steiner2025mcpnlweb} Steiner, A., Peeters, R., \& Bizer, C. (2025). \emph{MCP vs RAG vs NLWeb vs HTML: A Comparison of the Effectiveness and Efficiency of Different Agent Interfaces to the Web (Technical Report)}. arXiv:2511.23281. \url{https://arxiv.org/abs/2511.23281}.

\bibitem{agentready2026} DeepanshuPal (2026). \emph{agent-ready}. GitHub repository. Accessed 30 September 2026. \url{https://github.com/DeepanshuPal/agent-ready}.

\bibitem{chrome2026lighthouse} Chrome for Developers (2026). \emph{Lighthouse agentic browsing scoring}. Documentation, last updated 5 May 2026. \url{https://developer.chrome.com/docs/lighthouse/agentic-browsing/scoring}.

\bibitem{ballerine2026} Ballerine (2026). \emph{Ballerine Launches Trusted Agentic Commerce Governance Platform}. Press release, 15 January 2026. \url{https://www.prnewswire.com/news-releases/ballerine-launches-trusted-agentic-commerce-governance-platform-302662193.html}.

\bibitem{isagentready2026} IsAgentReady (2026). \emph{About IsAgentReady}. Methodology dated 2 September 2026. \url{https://isagentready.com/en/about}.

\bibitem{elogic2026} Okhrem, P. (2026). \emph{The Agentic Commerce Readiness Index 2026}. Elogic Commerce, 12 May 2026, updated 14 September 2026. \url{https://elogic.co/blog/agentic-commerce-readiness-index/}.

\bibitem{elnaffar2026agentready} Elnaffar, S., \& Rashidi, F. (2026). \emph{Designing Agent-Ready Websites for AI Web Agents: A Framework for Machine Readability, Actionability, and Decision Reliability}. arXiv:2607.12056. \url{https://arxiv.org/abs/2607.12056}.

\bibitem{peng2026acfc} Peng, Z. (2026). \emph{Agent-Callable Feature Coverage: Measuring Software Readiness for AI Agents}. arXiv:2609.35789. \url{https://arxiv.org/abs/2609.35789}.

\bibitem{jordan2026radar} Jordan, L., Peixoto, T. C., \& Ramos-Maqueda, M. (2026). \emph{RADAR: Readiness for AI Discovery and Agentic Reach}. arXiv:2609.28480. \url{https://arxiv.org/abs/2609.28480}.

\bibitem{wadi2026shopping} Wadi, D., \& Ma, Y. (2026). \emph{Shopping by algorithm: How agentic AI deploys human heuristics as a surrogate consumer}. arXiv:2609.28372. \url{https://arxiv.org/abs/2609.28372}.

\bibitem{gao2026bounds} Gao, S., \& Zhou, L. (2026). \emph{Can Agent Benchmarks Support Their Scores? Evidence-Supported Bounds for Interactive-Agent Evaluation}. arXiv:2605.10448. \url{https://arxiv.org/abs/2605.10448}.

\bibitem{manski2003partial} Manski, C. F. (2003). \emph{Partial Identification of Probability Distributions}. Springer Series in Statistics. Springer, New York. \url{https://doi.org/10.1007/b97478}.

\bibitem{horowitz2000missing} Horowitz, J. L., \& Manski, C. F. (2000). Nonparametric Analysis of Randomized Experiments with Missing Covariate and Outcome Data. \emph{Journal of the American Statistical Association}, 95(449), 77--84. \url{https://doi.org/10.1080/01621459.2000.10473902}.

\bibitem{ap2spec2026} Agent Payments Protocol (2026). \emph{AP2 specification, version 0.2}. Accessed 30 September 2026. \url{https://ap2-protocol.org/ap2/specification/}.

\bibitem{acp2026} OpenAI \& Stripe (2026). \emph{Agentic Commerce Protocol}. Specification version 2026-04-17. \url{https://github.com/agentic-commerce-protocol/agentic-commerce-protocol}.

\bibitem{ucp2026} Universal Commerce Protocol (2026). \emph{Specification overview}. Accessed 30 September 2026. \url{https://ucp.dev/specification/overview/}.

\bibitem{visa_tap2025} Visa (2025). \emph{Visa Introduces Trusted Agent Protocol: An Ecosystem-Led Framework for AI Commerce}. Press release, 14 October 2025. \url{https://usa.visa.com/about-visa/newsroom/press-releases.releaseId.21716.html}.

\bibitem{ftc2025fees} Federal Trade Commission (2025). Trade Regulation Rule on Unfair or Deceptive Fees. 16 CFR Part 464; 90 Fed.\ Reg.\ 2066 (10 January 2025), effective 12 May 2025. \url{https://www.ecfr.gov/current/title-16/part-464}.

\bibitem{blake2021salience} Blake, T., Moshary, S., Sweeney, K., \& Tadelis, S. (2021). Price Salience and Product Choice. \emph{Marketing Science}, 40(4), 619--636. \url{https://doi.org/10.1287/mksc.2020.1261}.

\bibitem{santana2020drip} Santana, S., Dallas, S. K., \& Morwitz, V. G. (2020). Consumer Reactions to Drip Pricing. \emph{Marketing Science}, 39(1), 188--210. \url{https://doi.org/10.1287/mksc.2019.1207}.

\bibitem{rfc9309} Koster, M., Illyes, G., Zeller, H., \& Sassman, L. (2022). \emph{Robots Exclusion Protocol}. RFC 9309, Internet Engineering Task Force. \url{https://www.rfc-editor.org/rfc/rfc9309.html}.

\bibitem{wilson1927} Wilson, E. B. (1927). Probable Inference, the Law of Succession, and Statistical Inference. \emph{Journal of the American Statistical Association}, 22(158), 209--212. \url{https://doi.org/10.1080/01621459.1927.10502953}.

\bibitem{bean2025construct} Bean, A. M., Kearns, R. O., Romanou, A., et al. (2025). \emph{Measuring what Matters: Construct Validity in Large Language Model Benchmarks}. NeurIPS 2025 Datasets and Benchmarks Track. arXiv:2511.04703. \url{https://arxiv.org/abs/2511.04703}.

\bibitem{cohen1960kappa} Cohen, J. (1960). A Coefficient of Agreement for Nominal Scales. \emph{Educational and Psychological Measurement}, 20(1), 37--46. \url{https://doi.org/10.1177/001316446002000104}.

\bibitem{menlo2012} Bailey, M., Dittrich, D., Kenneally, E., \& Maughan, D. (2012). The Menlo Report. \emph{IEEE Security \& Privacy}, 10(2), 71--75. \url{https://doi.org/10.1109/MSP.2012.52}.

\end{thebibliography}
\end{document}